%% file: main.tex
\documentclass[10pt,twocolumn,letterpaper]{article}

\usepackage[pagenumbers]{cvpr} 

\definecolor{cvprblue}{rgb}{0.21,0.49,0.74}
\usepackage[pagebackref,breaklinks,colorlinks,allcolors=cvprblue]{hyperref}
\usepackage{microtype}
\usepackage{url}            
\usepackage{booktabs}       
\usepackage{amsfonts}       
\usepackage{nicefrac}       
\usepackage{microtype}      
\usepackage{xcolor}         

\usepackage{graphicx}
\usepackage{algorithm,multicol}
\usepackage[noend]{algpseudocode}
\usepackage{amsmath}
\usepackage{xcolor}
\usepackage{amsthm, bm}

\usepackage{multirow}
\usepackage{graphicx}
\usepackage{float}
\usepackage{makecell}
\usepackage{bbold}
\usepackage{colortbl}
\usepackage{subcaption}
\usepackage{cuted}
\usepackage{capt-of}
\usepackage{pifont}
\usepackage{soul}
\newcommand{\cmark}{\ding{51}}%
\newcommand{\xmark}{\ding{55}}%

\def\confName{CVPR}
\def\confYear{2025}
\def\CMeRT{CMeRT}

\title{Context-Enhanced Memory-Refined Transformer for Online Action Detection}

\author{Zhanzhong Pang\\
National University of Singapore\\
{\tt\small pang@comp.nus.edu.sg}
\and
Fadime Sener\\
Meta Reality Labs\\
{\tt\small famesener@meta.com}
\and
Angela Yao\\
National University of Singapore\\\
{\tt\small ayao@comp.nus.edu.sg}
}

\begin{document}
\maketitle
\input{sec/0_abstract}    
\input{sec/1_intro}
\input{sec/2_relate_work}
\input{sec/3_pre}
\input{sec/4_method}

\input{sec/5_experiment}
\input{sec/6_conclusion}

\vspace{1.5mm}
\noindent \textbf{Acknowledgment.}
This research is supported by the National Research Foundation, Singapore and DSO National Laboratories under its AI Singapore Programme (AISG Award No: AISG2-RP-2020-016). Any opinions, findings and conclusions or recommendations expressed in this material are those of the author(s) and do not reflect the views of National Research Foundation, Singapore.

{
    \small
    \bibliographystyle{ieeenat_fullname}
    \bibliography{main}
}

\clearpage
\input{sec/X_suppl}

\end{document}

%% file: sec/0_abstract.tex
\begin{abstract}

Online Action Detection (OAD) detects actions in streaming videos using past observations. State-of-the-art OAD approaches model past observations and their interactions with an anticipated future. The past is encoded using short- and long-term memories to capture immediate and long-range dependencies, while anticipation compensates for missing future context. We identify a training-inference discrepancy in existing OAD methods that hinders learning effectiveness. The training uses varying lengths of short-term memory, while inference relies on a full-length short-term memory. As a remedy, we propose a Context-enhanced Memory-Refined Transformer (\CMeRT).  \CMeRT\ introduces a context-enhanced encoder to improve frame representations using additional near-past context. It also features a memory-refined decoder to leverage near-future generation to enhance performance. \CMeRT~\footnote{Code: \url{https://github.com/pangzhan27/CMeRT}.} achieves state-of-the-art in online detection and anticipation on 
THUMOS'14, CrossTask, and EPIC-Kitchens-100. 
\end{abstract}

%% file: sec/1_intro.tex
\vspace{-6mm}
\section{Introduction}
\label{sec:intro}

\begin{figure}[t]
\centering
\subfloat[][Imbalanced context exposure in short-term memory, where frames can only access the past: current frame $t$ has full short-term context ($t_s \backsim t$), while intermediate frame $t_m$ has only partial ($t_s \backsim t_m$).]{\includegraphics[scale=0.165]{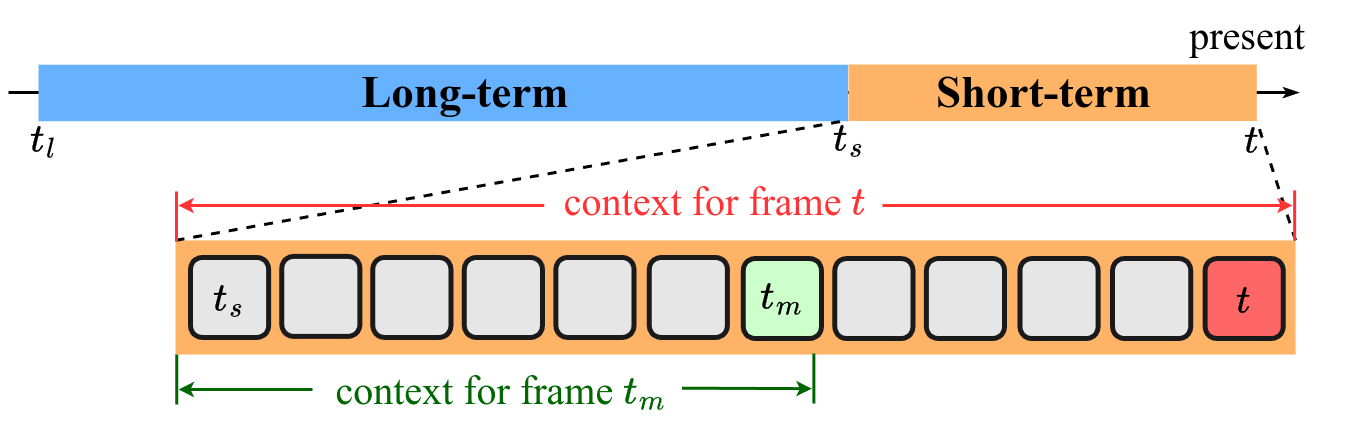}} \\
\vspace{1mm}
\subfloat[][Frame losses within the short-term memory, showing learning biases toward earlier and intermediate frames.]{\includegraphics[scale=0.52]{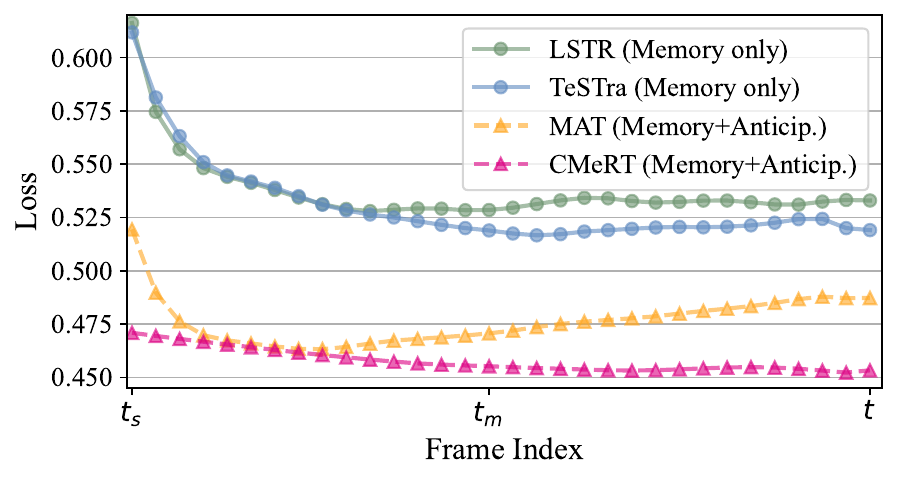} \vspace{-2mm}} \\
\vspace{-3mm}
\caption{Existing methods exhibit poorly learned frame representations due to imbalanced context exposure and non-causal leakage.}
\label{fig:motivation} 
\vspace{-7mm}
\end{figure}

Online Action Detection (OAD) identifies actions in a video stream based on only past observations. This task is crucial for applications like autonomous driving~\cite{kim2019grounding}, surveillance~\cite{shu2015joint,pang2020self}, and AR assistants~\cite{ding2025spatial,koppula2015anticipating,zhou2018towards,sener2022transferring,sener2022assembly101,ben2021ikea}, where immediate and accurate detection is essential.

Effective OAD requires sufficient temporal support from past frames. Recent methods~\cite{xu2021long,zhao2022real,wang2023memory, guermal2024joadaa, chen2022gatehub} partition past observations into long- and short-term memories, and enrich frames in short-term memory with both immediate and long-range contexts to capture a range of temporal dependencies. A causal mask~\cite{vaswani2017attention} is typically applied to short-term frames, restricting each frame to access only previous frames so that predictions rely solely on past. This causal masking makes each frame in short-term memory function as if it were the most recent observed frame, allowing all short-term frames to be used as training samples to improve training efficiency. State-of-the-art OAD approaches~\cite{wang2023memory, guermal2024joadaa, chen2022gatehub} further leverage action anticipation to generate pseudo-futures, compensating for the absence of a true future context in OAD. In these works, all frames in the short-term memory are used for training, with each frame having different level of immediate context. However, inference relies exclusively on the \textit{latest frame}, which has access to the full short-term context.

In this paper, we identify a training-inference discrepancy present in state-of-the-art OAD methods and reveal how this discrepancy introduces biases into training, limiting effective modeling of the \textit{latest frame} for inference. As shown in \cref{fig:motivation}, we observe two sources of bias. (1) The causal mask applied to the short-term memory exposes frames to imbalanced amounts of context relative to frame position. While the latest frame $t$ has full short-term context, the earliest $t_s$ has none due to its position. This imbalance in context degrades learning quality, resulting in less informative representations (large loss) for earlier frames in all SOTA models~\cite{xu2021long, zhao2022real, wang2023memory}. These poor representations hinder the classifier's ability to effectively predict the latest frame. (2) Using anticipation as the pseudo-future to enrich short-term memory introduces learning biases across short-term frames. Since anticipated context is derived from the full short-term memory, early frames indirectly access their future short-term frames through interaction with anticipation, creating non-causal leakage. This leakage skews training by favoring intermediate frames, which have access to both past and future context. MAT~\cite{wang2023memory} exhibits a valley-shaped loss curve, indicating a learning bias toward intermediate frames, which harms the training and inference of the latest frame. 

Based on these key findings, we revisit memory-based OAD methods and focus on mitigating the observed training-inference discrepancy. Building upon the long- and short-term formulation, we introduce the \textbf{C}ontext-enhanced \textbf{Me}mory-\textbf{R}efined \textbf{T}ransformer (\CMeRT). \CMeRT\ incorporates a context-enhanced module to supplement the context for earlier frames in short-term memory, improving training and yielding better frame representations. In addition, \CMeRT\ includes a memory-refinement module that enhances short-term memory using generated near-future frames. Unlike \cite{wang2023memory}, our anticipated future is derived from long-term memory, preventing non-causal leakage and reducing the learning bias toward intermediate frames. 

We also present new protocols for OAD, highlighting several weaknesses in the OAD literature, such as outdated features, limited evaluation metrics, and constrained datasets, which collectively hinder advancements in the field. We evaluate all state-of-the-art OAD methods using stronger visual features~\cite{oquab2023dinov2}, implement event-based metrics for better action-level performance assessment, and provide extra comparisons on CrossTasks, a procedural activity dataset where modeling long-term dependencies is essential. 

Our contribution are
\begin{itemize}
\item revealing a training-inference gap of current OAD approaches, caused by naive memory and anticipation processing.
\item proposing a new OAD architecture \CMeRT\ with improved memory and anticipation formulations to mitigate the training-inference discrepancy.
\item presenting a new OAD benchmark, along with new protocols to update the data, features, and metrics of OAD.
\item achieving SOTA detection and anticipation performance on three challenging datasets, under both the standard and our newly proposed protocols.
\end{itemize}

%% file: sec/2_relate_work.tex
\section{Related Work}
\label{sec:related}

\textbf{Online Action Detection and Anticipation.} 
Early works focus on effectively leveraging past information to model long-range temporal interactions. RNN-based methods~\cite{an2023miniroad, de2018modeling, eun2020learning, gao2017red, li2016online, xu2019temporal} model sequences recurrently but struggle to capture long-range dependencies. Techniques like two-stream networks~\cite{de2018modeling}, IDN~\cite{eun2020learning}, and GatedHub ~\cite{chen2022gatehub} improve temporal modeling, while approaches like \cite{shou2018online, gao2019startnet, gao2017red} decompose the tasks into action recognition and action start point detection. Some works integrate online detection and anticipation, leveraging a predicted future to improve action detection. These methods employ RNN cells~\cite{xu2019temporal,kim2021temporally} or Transformers~\cite{wang2021oadtr,wang2023memory} for future anticipation.

\noindent \textbf{Transformers for Online Action Detection.} Transformers have shown great success in vision~\cite{alexey2020image, liu2021swin, wang2021pyramid} and video tasks~\cite{liu2022end, chen2023videollm, arnab2021vivit, zhang2022actionformer}. Recently, LSTR~\cite{xu2021long} and TeSTra~\cite{zhao2022real} explore long- and short-term memories using transformers for OAD. These approaches partition the entire history into long- and short-term memories, and use transformers for long-term compression~\cite{xu2021long, li2023blip}, short-long term interactions, future action anticipation~\cite{guermal2024joadaa, wang2023memory}. However, these memory-based methods fail to consistently model frames in short-term memory and introduce non-causal leakage when using anticipation results to enhance detection. 

In this paper, we systematically analyze short-term memory modeling and joint detection-anticipation, proposing an optimal transformer-based solution. We also advance OAD research using SOTA features, more representative metrics, and a new benchmark on a procedural activity dataset, opening avenues for OAD in new task settings~\cite{chatterjee2023opening, lin2023video} and applications~\cite{kukleva2024x,  pang2024cost, pang2024long}.

%% file: sec/3_pre.tex
\section{Diagnosing Context Modeling for OAD}

\label{sec: pre}
\subsection{Preliminaries}

\textbf{Online Action Detection} identifies the ongoing action at time $t$ in a video stream, relying solely on observations up to and including $t$. Formally, given a video stream up to $t$ as $V = \{v_0, v_1, \cdots, v_t\}$, the objective is to predict the action, $y_t \in \{0, 1, 2, \cdots, C\}$, where $0$ represents a background class, and $y_t$ is the action in frame $t$. 

\begin{figure}[tb]
 \centering
 \vspace{-3mm}
 \centering
 \includegraphics[width=1.0\linewidth]{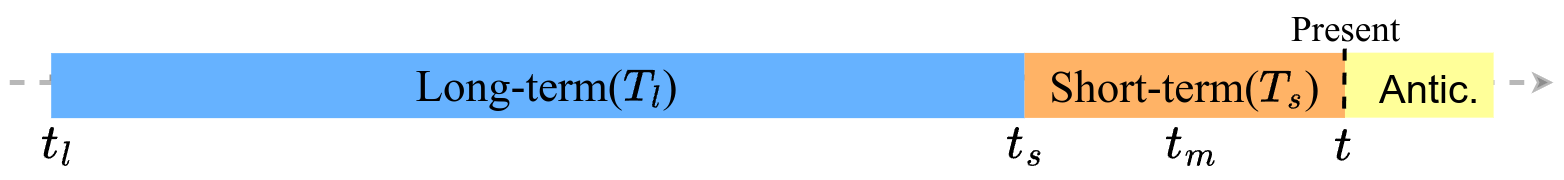}
 \small
 \centering
 \resizebox{1.0\columnwidth}{!}{
 \begin{tabular}{c|c|c|c|c}
 \hline
 \multirow{2}{*}{Phase} & \multirow{2}{*}{Timestamp} &\multicolumn{3}{c}{Encoded context} \\
 \cline{3-5}
 & & Long & Short & Antic.\\
 \hline
 \multirow{3}{*}{Train}& $t_s$ & \multirow{4}{*}{\color{red}{$\overline{t_l \backsim t_s}$}} & $t_s$ & \multirow{4}{*}{\makecell{\color{red}{$\overline{t_l \backsim t_s}$} \color{black}{,} \ \color{red}{$\overline{t_s \backsim t}$} \color{black}{,}  \\ \color{blue}{$\widehat{t \backsim t+\tau}$}}} \\
 \cline{2-2} \cline{4-4}
 & $t_m$ & & $t_s \backsim t_m$ & \\
 \cline{2-2} \cline{4-4}
 & $t$ & & $t_s \backsim t$ & \\
 \cline{1-2} \cline{4-4}
 Inference & $t$ & & $t_s \backsim t$ & \\
 \hline
 \end{tabular}}
 \vspace{-1mm}
 \caption{Context analysis for short-term frames. {\color{red}{$\overline{\ \ast \ }$}} for indirectly encoded, {\color{blue}{$\widehat{\ \ast \ }$}} for generated, and the rest for direct context. }
 \label{tab:ana_ctx} 
 \vspace{-5mm}
\end{figure}

Recent memory-based methods LSTR~\cite{xu2021long} and Testra~\cite{zhao2022real} employ transformers. They operate on pre-extracted frame-wise features\footnote{Feature extractors typically operate on short segments of consecutive frames. We abuse the term ``frame'' as segment-wise features for simplicity.}, $f_t \in \mathbf{R}^D$, and partition the history into short, $M_S = \{f_i\}_{i=t_s}^t$, and long-term memories, $M_L = \{f_i\}_{i=t_l}^{t_s-1}$, to capture both immediate and long-range dependencies. The memory bounds are defined by $t_s = t-T_s+1, t_l = t-T_l-T_s+1$ where $T_s$ and $T_l$ denote the lengths of short- and long-term, respectively (see \cref{tab:ana_ctx} ). The long-term memory is compressed into a latent representation using transformers with learnable queries~\cite{xu2021long, li2023blip, guermal2024joadaa}. 

To enhance training efficiency, all frames in the short-term memory are used as training samples, leveraging the same precomputed short- and long-term memories (computed only once). To ensure causality, a mask is applied to the short-term memory, restricting each frame to access only previous frames and make predictions as if it were the most recent observed frame. The short-term frames are enriched by interacting with both their immediate past frames from the short-term memory and the compressed long-term memory.

In the memory-based setting, training involves sampling a ``current" timestamp $t$, with long and short-term memories constructed by padding or cropping past observations up to $t$. $t$ is sampled either by (1) a sliding window with a random start time and stride equal to the short-term length, $T_s$, and (2) event-based sampling, where $t$ is randomly chosen within the duration of a non-background action. In both cases, all frames in a  short-term memory serve as training samples to enhance efficiency. During inference, a sliding window with stride 1 and fixed start time at 0 simulates an online streaming setup, predicting one new frame at a time.

\noindent \textbf{Online Action Anticipation} predicts the future action occurring after an interval, $\tau$, \ie $a_{t +\tau} \in \{0, 1, 2, \cdots, C\}$, where $a_{t +\tau}$ is the anticipated action in frame $t+\tau$, based on visual evidence up to $t$. Online action detection is a special case with $\tau=0$. Thus, recent approaches~\cite{wang2023memory, guermal2024joadaa} integrate detection and anticipation within a unified network and utilize anticipation outputs as pseudo-future to improve detection.

\subsection{Long- and Short-term Memories}
\label{subsec: rep}
Since all frames in a short-term memory serve as training samples, we first analyze the context accessible to each frame. We use $t_s$, $t_m$ and $t$, representing the start, middle, and end of the short-term memory. For \textbf{short-term context}, frames $t_s$, $t_m$ and $t$ access varying amounts of short-term history due to the causal masking (\cref{tab:ana_ctx}). Frame $t$ has access to the entire short-term memory from $t_s$ to $t$, while $t_s$ accesses only itself due to its start position, and $t_m$ accesses frames from $t_s$ to $t_m$. For \textbf{long-term context}, the long-term memory $M_L$ is compressed into a latent representation $\widehat{M_L}$. As a result, all three frames $t_s$, $t_m$ and $t$ can only access the long-term $M_L$ indirectly through $\widehat{M_L}$. This compression leads to loss of finer details, preventing frame $t_s$ from recovering its immediate past, despite interacting with $\widehat{M_L}$.

We observe that short-term frames exposed to varying contexts exhibit different learning behaviors. As shown in \cref{fig:motivation} (b), loss curves of short-term memory for existing memory-based models~\cite{xu2021long, zhao2022real, wang2023memory} reveal that earlier frames, like $t_s$ and those immediately following, incur higher loss due to insufficient immediate context, resulting in less informative representations. Despite this, these models treat all short-term frames equally during training, with poorly represented frames being low-quality samples. These samples ultimately impair the classifier's ability to effectively predict the latest frame, which is the focus during inference.

Another observation is that for methods that do not use anticipation, such as LSTR~\cite{xu2021long} and Testra~\cite{zhao2022real}, their loss curves in \cref{fig:motivation} (b), show that extra context beyond the immediate history offers limited benefits. For example, frames $t_m$ and $t$ both have access to an immediate past of at least $T_s/2$ frames. Yet the extra context available to frame $t$ preceding $t_m$ provides no benefit, suggesting that excessive immediate context may not be necessary.

\begin{figure}
 \centering
 \includegraphics[width=0.8\linewidth]{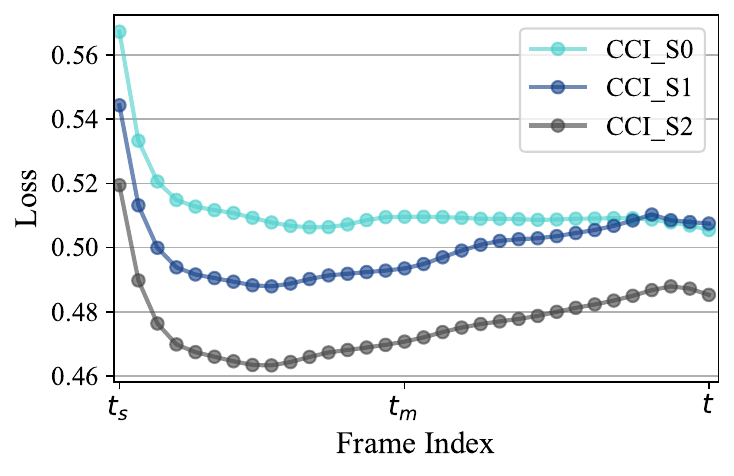}
 \vspace{-4mm}
 \caption{Frame losses within the short-term memory in MAT~\cite{wang2023memory} at different rounds of accessing anticipated future.}
 \label{fig:cci_loss}
 \vspace{-6mm}
\end{figure}

\subsection{Pseudo-Future Context}
\label{subsec: leak}
Building on LSTR~\cite{xu2021long}, MAT~\cite{wang2023memory} unifies anticipation and detection by using anticipation to generate a pseudo-future for frames $t+1 \backsim t+\tau$, to improve detection. MAT introduces Conditional Circular Interaction(CCI) to enable iterative interaction between short-term and anticipation.  However, we uncover that its CCI allows short-term frames to indirectly access subsequent frames, compromising the causal nature of OAD and also leading to less informative representations for the later frames. 

As shown in~\cref{tab:ana_ctx}, anticipation output is generated based on compressed long and short-term memory via cross-attention. This indirectly incorporates information from the entire short-term memory from $t_s$ to $t$. When updating the short-term memory with anticipation, earlier frames, such as $t_m$, indirectly access subsequent frames ($t_m$ to $t$) through queries to the anticipated future, resulting in non-causal leakage. As such, MAT exhibits higher losses for current frames than intermediate ones, suggesting a learning bias towards intermediate frames, as shown in~\cref{fig:motivation}. This bias is further confirmed in \cref{fig:cci_loss}, where learning initially focuses on the current frame without interacting with the anticipated future~(CCI\_S0), then shifts towards intermediate frames after several rounds of accessing anticipation~(CCI\_S2).

%% file: sec/4_method.tex
\section{\CMeRT} 
\label{sec:method}

\begin{figure*}[th!]
 \centering
 \includegraphics[width=1\linewidth]{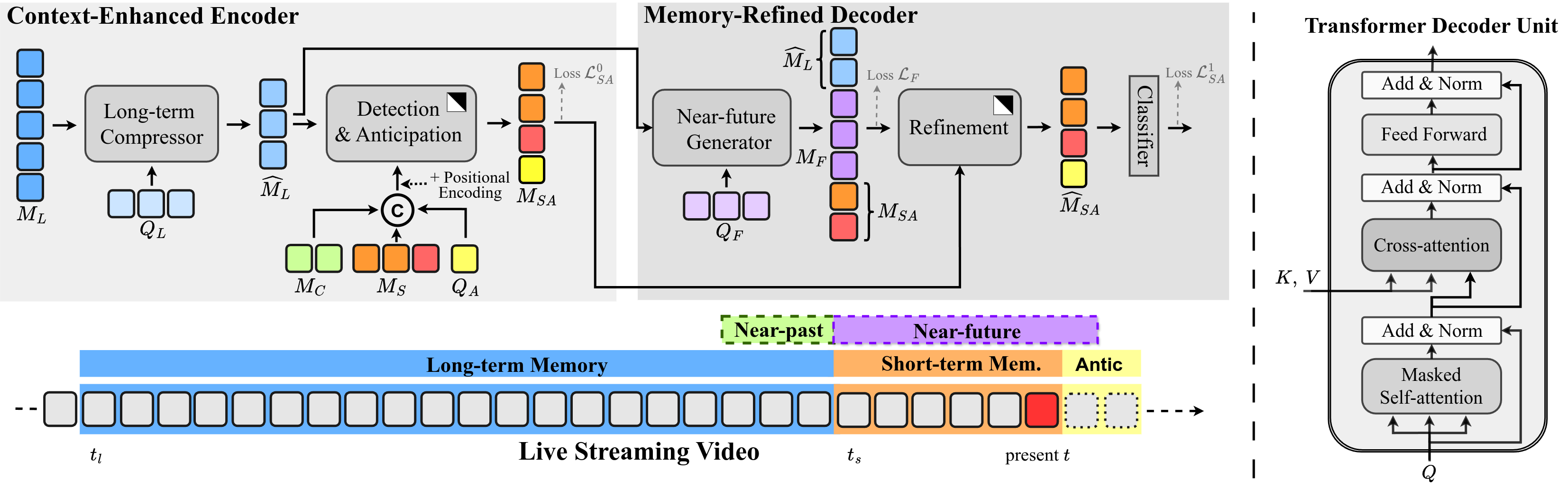}
 \vspace{-0.8cm}
 \caption{Framework of Context-Enhanced Memory-Refined Transformer. The model is in an encoder-decoder formulation, operating on five context partitions: long-term, short-term, anticipation, near-past, and near-future. The Context-Enhanced Encoder compresses the long-term memory $M_L$ and encodes the short-term memory with anticipation as $M_{SA}$ using the compressed long-term $\widehat{M_L}$ and near-past context $M_C$. The Memory-Refined Decoder generates the near-future context $M_F$ from $\widehat{M_L}$ and refines $M_{SA}$ using $M_F$. A weight-shared classifier is adopted to classify both short-term $M_{SA}$ and $\widehat{M_{SA}}$ and near-future $M_F$. All modules are build upon Transformer Decoder Unit.}
 \label{fig: framework}
 \vspace{-0.5cm}
\end{figure*}

We propose Context-enhanced Memory-Refined Transformer (\CMeRT), a unified framework for detection and anticipation(see \cref{fig: framework}) that addresses the imbalanced contexts and learning biases outlined in \cref{sec: pre}. Building upon \cite{xu2021long, zhao2022real}, we maintain three non-overlapping context partitions: long-term memory, short-term memory, and anticipated context. In addition, we introduce near-past and near-future contexts to mitigate the training-inference discrepancy caused by biases in frame representation learning. \CMeRT\ comprises a Context-Enhanced Encoder(\cref{subsec: encoder}) and a Memory-Refined Decoder(\cref{subsec: decoder}). The encoder leverages near-past context to learn consistent short-term representations, while the decoder queries generated near-future context with these learned short-term representations for decoding. 
The transformer decoder unit~\cite{vaswani2017attention} serves as the basic building block (\cref{subsec: tdu}) of \CMeRT.

\subsection{Transformer Decoder Unit}
\label{subsec: tdu}
Our model uses a Transformer Decoder Unit (TDU) as the core building block, as illustrated in \cref{fig: framework}. The TDU takes in queries $Q$, keys $K$, values $V$, and a mask $G$ as input and outputs updated queries $\widehat{Q}$ based on keys and values.
\vspace{-1mm}
\begin{equation}
 \widehat{Q} = \text{TDU}(Q, K, V, G)
 \label{eq:tdu}
 \vspace{-1mm}
\end{equation}

The TDU consists of a self-attention module for query interactions, a cross-attention module for query-key-value interactions, and a feed-forward network (FFN) for processing the attended information. Each module is followed by a skip connection and LayerNorm~\cite{vaswani2017attention}.
\vspace{-1mm}
\begin{align}
 Q' &= \text{Self-Attention}(Q, \ G) + Q \nonumber\\
 Q'' &= \text{Cross-Attention}(Q', \ K, \ V) + Q' \\
 \widehat{Q} &= \text{FFN}(Q'') + Q'' \nonumber 
 \label{eq: tdu_detail}
 \vspace{-1mm}
\end{align}

\subsection{Context-Enhanced Encoder}
\label{subsec: encoder}
As discussed in \cref{subsec: rep}, earlier short-term frames lack immediate context, resulting in poor representations. The Context-Enhanced Encoder addresses this by incorporating near-past context, improving early frames' representations and overall learning effectiveness.
Our encoder comprises a long-term compressor and a detection \& anticipation module. It compresses long-term memory to improve efficiency, then encodes short-term memory based on the compressed long-term and the near-past context. The encoded short-term memory serves as the queries to the Memory-Refined Decoder, rather than being the final output. 

\noindent \textbf{Long-term Compression. }
Given video features up to time $t$ as $F = \{f_i\}_{i=0}^t$, long-term memory stores features $M_L = \{f_i\}_{i=t_l}^{t_s-1}$. For computational efficiency, $M_L$ is first uniformly sub-sampled at a coarse temporal scale. Following prior work~\cite{xu2021long}, a two-stage compression module is applied to $M_L$ to generate an abstract representation $\widehat{M}_L$. Each stage uses a TDU with learnable queries $Q_L$ for compression. Positional encoding is omitted since the compression process removes temporal information.
\vspace{-1mm}
\begin{align}
 M_L' &= \text{TDU}(Q_L^0, \ M_L, \ M_L, \ \text{None}) \\
 \widehat{M}_L &= \text{TDU}(Q_L^1, \ M_L', \ M_L', \ \text{None}) \nonumber
 \label{eq: compress}
 \vspace{-6mm}
\end{align}
where "None" indicates that no mask is applied. 

\noindent \textbf{Detection \& Anticipation. }
Our Detection \& Anticipation module employs a TDU with masked self-attention~\cite{xu2021long, wang2023memory}. Here, a causal mask, $G$, should be applied to the short-term memory to ensure each frame only accesses preceding frames. This causal masking, however, causes an imbalance in immediate context across frames~(see \cref{fig:motivation} and \cref{tab:ana_ctx}). To address this, we extract near-past memory, $M_C = \{f_i\}_{i=t_s - T_c}^{t_s -1}$ and append it to the short-term memory to provide extra immediate context for earlier frames at time $t_s$. Here, $T_c$ represents the near-past context length and $T_c \ll T_l$. While the near-past context, $M_C$, and the long-term memory, $M_L$, overlap, the compression applied to the long-term memory $M_L$ loses fine-grained details, which are retained in the extended short-term memory $M_C$. For anticipation, learnable queries $Q_A$ of length $T_a$ are used. Finally, $M_C$, $M_S$ and $Q_A$ are concatenated and passed to the TDU, with sinusoidal positional encoding~\cite{vaswani2017attention} added to preserve their temporal structure. 

Overall, short-term frames are updated through masked self-attention with their immediate past ($M_c$ and $M_s$), as well as cross-attention with the long-range past ( $\widehat{M_L}$). After contributing additional context, the updated near-past memory is discarded from the TDU output, yielding $M_{SA}$ as the anticipated, updated short-term memory. 
\vspace{-1mm}
\begin{align}
 M_{SA} = \text{TDU}(&M_C || M_S || Q_A , \ \widehat{M}_L || M_S || Q_A, \\
 & \widehat{M}_L || M_S || Q_A, \ G \ )_{[T_c:T_c+T_s +T_a ]} \nonumber
 \label{eq:det}
\end{align}
\vspace{-1mm}
\noindent where $||$ represents concatenation.

\subsection{Memory-Refined Decoder}
\label{subsec: decoder} 
Previous works~\cite{wang2023memory, guermal2024joadaa} have highlighted the benefits of using anticipation in detection; however, they introduce learning biases when utilizing pseudo-future information. To address this, we design a near-future generator in the decoder that is independent of short-term memory. Our decoder also incorporates a memory refinement module that refines the encoded short-term memory from the Context-Enhanced encoder using the generated near-future context.

\noindent\textbf{Near-Future Generator. }
In \cref{subsec: leak}, we demonstrated that near-future information boosts intermediate frames' performance through non-causal leakage. However, current method~\cite{wang2023memory} provides near-future~($t+1 \backsim t+\tau$) via anticipation only for the current frame $t$, not earlier ones. To address this, we propose the Near-Future Generator, which generates near-future information for all short-term frames. 
In addition, we avoid using short-term memory and instead leverage compressed long-term to generate near-future for all frames in short-term. The use of compressed long-term eliminates the non-causal leakage outlined in \cref{subsec: leak}.  
Specifically, a TDU with learnable queries $Q_F$ of length $T_f$ retrieves useful information from compressed long-term $\widehat{M}_L$. The generated future $M_F$ spans from $t_s$ to $t_s +T_f$, providing near-future information for all short-term frames.
\begin{equation}
 M_{F} = \text{TDU}(Q_F, \widehat{M}_L, \widehat{M}_L, \text{None}).
 \label{eq:fut}
\end{equation}

\noindent\textbf{Memory Refinement. }
We refine the encoded short-term and anticipation $M_{SA}$ from the Context-Enhanced encoder with the generated near-future $M_F$. The memory refinement module takes $M_{SA}$ as queries and the fusion of long $\widehat{M}_L$, short $M_{SA}$ and near-future $M_F$ as keys and values for memory refinement. Since $M_F$ is generated based upon the compressed long-term $\widehat{M}_L$, using it to update $M_{SA}$ avoids contamination from indirectly accessing subsequent frames. The refined memory $\widehat{M}_{SA}$ is then passed to a classifier for action detection and anticipation.
\vspace{-1mm}
\begin{align}
 \widehat{M}_{SA} = \text{TDU}(M_{SA}, \ \widehat{M}_L || M_{SA} || M_F, \ \widehat{M}_L || M_{SA} || M_F, \ G).
 \label{eq:refine}
\end{align}

\vspace{-4mm} 
\subsection{Training and Inference}
Following prior works~\cite{xu2021long, zhao2022real, wang2023memory}, training samples are generated using a sliding window~(THUMOS'14 and CrossTask) with a random start and stride $T_s$, or event-based sampling~(EK100). The Context-Enhanced Encoder $M_{SA}$ and Memory-Refined Decoder $\widehat{M}_{SA}$ predictions are fed into a shared classifier, yielding action probabilities $P_{SA}$ and $\widehat{P}_{SA}$. Cross-entropy loss is applied to the entire short-term memory and the anticipation based on ground truth labels $Y_{SA}$:
\vspace{-4mm}
\begin{align}
 \mathcal{L}_{SA}^0 = -\sum_{i=1}^{T_s + T_a} Y_{SA}^i \log P_{SA}^i, \ \mathcal{L}_{SA}^1 = -\sum_{i=1}^{T_s + T_a} Y_{SA}^i \log \widehat{P}_{SA}^i
 \label{eq: loss_det}
\end{align}

\vspace{-3mm}
\noindent Additionally, the generated future features, $M_F$, are passed through the same classifier, yielding the probability, $P_F$, based on the future targets, $Y_F$:
\vspace{-2mm}
\begin{align}
 \mathcal{L}_{F} = -\sum_{i=1}^{T_f} Y_{F}^i \log P_{F}^i
 \label{eq: loss_fut}
 \vspace{-5mm}
\end{align}

\vspace{-2mm}
\noindent The final training loss is defined as
\begin{align}
 \mathcal{L} = \mathcal{L}_{SA}^1 + \lambda_1 \mathcal{L}_{SA}^0 + \lambda_2 \mathcal{L}_{F} 
 \label{eq: loss}
 \vspace{-5mm}
\end{align}
where $\lambda_1$ and $\lambda_2$ are the balancing coefficients.

During inference, samples are generated with a sliding window (stride 1, start time 0) to simulate online streaming setup. Detection and anticipation are inferred simultaneously from $\widehat{P}_{SA}$ of the Memory-Refinement module, with outputs split as $\widehat{P}_S = {\widehat{P}_{SA[:T_s]}}$, and $\widehat{P}_A = {\widehat{P}_{AS[T_s:T_s+T_a]}}$. For detection, only the last frame prediction in $\widehat{P}_{S}$ is used for action detection. For anticipation, the corresponding frame in $\widehat{P}_A$ is selected based on the time gap $\tau$ for forecasting.

%% file: sec/5_experiment.tex
\section{Experiments}
\label{sec:experiments}

\subsection{Dataset, Evaluation, and Implementation}
\noindent \textbf{Dataset.} 
We experiment on three datasets: THUMOS'14(TH'14)~\cite{idrees2017thumos}, EPIC-Kitchens-100~(EK100)~\cite{damen2022rescaling}, and CrossTask(CT)~\cite{zhukov2019cross}. Each dataset has unique characteristics. THUMOS'14 is sparsely annotated, with most videos containing a single action; EK100 contains fine-grained actions where long-term dependencies are less critical; CrossTask features procedural videos with strong temporal action relationships. THUMOS'14 includes 413 untrimmed sports videos annotated with 20 classes. Following ~\cite{xu2019temporal, xu2021long, wang2023memory}, we train on the validation set (200 videos) and evaluate on the test set (213 videos). EK100 contains 100 hours of egocentric kitchen videos, labeled with 97 verb classes, 300 noun classes, and 3806 action classes. We adopt the train/val split from \cite{furnari2020rolling}. CrossTask contains 2750 videos of 18 primary tasks comprising 212 hours of video with 105 action classes. 

\noindent \textbf{Evaluation.} 
Though our work targets OAD, \CMeRT~effectively handles detection and anticipation in a unified way, so we report results for both detection and anticipation. Following~\cite{de2016online, xu2021long, zhao2022real, wang2023memory}, we evaluate online action detection and anticipation using per-frame mean average precision (mAP) for THUMOS'14 and CrossTask, and mean Top-5 Recall for verb/noun/action in EK100. For anticipation, we apply a period ranging from 0.25s to 2.0s with a stride of 0.25s for THUMOS'14, and a fixed interval of 1s for EK100~\cite{wang2023memory}.

\begin{table}[t]
\caption{Length of each partitioned context in seconds, including long-term, short-term, anticipation, near-past and near-future.}
\centering
\vspace{-2mm}
\small
\setlength{\tabcolsep}{1.3mm}
\begin{tabular}{cccccc}
\hline
Dataset & Long & Short & Anticipation & Near-past & Near-future \\ 
\hline
TH'14 & 256 & 4 & 2 & 0.5 & 12 \\ 
EK100 & 128 & 8 & 2 & 2 & 8 \\
CrossTask & 128 & 10 & 2 & 8 & 12 \\ 
\hline
\end{tabular}
\label{tab:mem_len}
\vspace{-5mm}
\end{table}

\noindent \textbf{Implementation.} 
We use pre-extracted frame features as in \cite{xu2021long, zhao2022real, wang2023memory}. For THUMOS'14, we employ two-stream features (at 4 FPS) with ResNet-50 for visuals and BN-Inception for motion~\cite{carreira2017quo, zhao2022real, xu2021long}. On EK100, we use features from a two-stream TSN (4 FPS) pretrained on ImageNet~\cite{furnari2020rolling, zhao2022real}. For CrossTask, we incorporate RGB I3D~\cite{carreira2017quo} features at 1 FPS with audio VGG features~\cite{zhukov2019cross}. \cref{tab:mem_len} provides the lengths of the partitioned long-term, short-term, anticipation, near-past, and near-future contexts. For the two-stage long-term compression, we set the number of queries to 16-32 for THUMOS'14 and CrossTask, and 16-16 for EK100. Balancing coefficients $\lambda_1$ and $\lambda_2$ in \cref{eq: loss} are 0.2 and 0.5, respectively. Similar to~\cite{wang2023memory, zhao2022real}, we use the Adam optimizer with weight decay and a cosine annealing schedule with warm-up. For EK100, we adopt equalization loss~\cite{tan2020equalization} to address the long-tail action, along with MixClip~\cite{zhao2022real} and MixClip++~\cite{wang2023memory} for data augmentation. Further details of hyperparameters are in Supplementary B.

\begin{table}[hbt!]
\centering
\vspace{-1mm}
\caption{OAD performance comparison, measured by mAP for THUMOS'14 and CrossTask, and mean Top-5 Recall for EK100. }
\vspace{-2mm}
\label{tab: oad}
\centering
\small
\setlength{\tabcolsep}{0.5mm}
\begin{tabular}{cccccc}
\hline
Method & TH'14 \\ 
\hline
LSTR~\cite{xu2021long} & 69.5 \\ 
GateHub~\cite{chen2022gatehub} & 70.7 \\
Testra~\cite{zhao2022real} & 71.2 \\
MAT~\cite{wang2023memory} & 71.6 \\ 
JOAAD~\cite{guermal2024joadaa} & 72.6 \\
MAT-rw & 71.7 \\
MAT-stream & 58.1 \\
\CMeRT\ & \textbf{73.2} \\
\hline
\end{tabular}
\ \ \
\setlength{\tabcolsep}{0.9mm}
\begin{tabular}{ccccc}
\hline
\multirow{2}{*}{Method} & \multirow{2}{*}{CT} & \multicolumn{3}{c}{EK100} \\ 
\cline{3-5} & & Verb & Noun & Action \\ \hline
LSTR~\cite{xu2021long} & 33.0 & 39.6 & 44.1 & 22.6 \\ 
Testra~\cite{zhao2022real} & 33.4 & 39.7 & 45.6 & 25.1 \\
MAT~\cite{wang2023memory} & 33.9 & 44.5 & \textbf{48.3} & 26.3 \\ 
MAT-rw & 34.1 & 46.3 & 47.3 & 26.7 \\ 
MAT-stream & 27.9 & 43.5 & 45.1 & 24.7 \\
\CMeRT\ & \textbf{35.9} & \textbf{47.1} & \textbf{48.3} & \textbf{27.6} \\
\hline
\end{tabular}
\vspace{-4mm}
\end{table}

\begin{table}[t]
\caption{Action anticipation results on THUMOS'14.}
\label{tab: aa_th}
\vspace{-2mm}
\small
\setlength{\tabcolsep}{0.9mm}
\begin{tabular}{cccccccccc}
\hline
\multirow{2}{*}{Method} & \multicolumn{8}{c}{mAP@$\tau$} & \multirow{2}{*}{Avg.}\\ 
\cline{2-9} & 0.25 & 0.50 & 0.75 & 1.0 & 1.25 & 1.50 & 1.75 & 2.0 & \\ \hline
IIU~\cite{lee2022learning} & 55.6 & 55.3 & 54.6 & 53.1 & 51.4 & 49.8 & 48.5 & 46.9 & 51.9 \\
LSTR~\cite{xu2021long} & 60.4 & 58.6 & 56.0 & 53.3 & 50.9 & 48.9 & 47.1 & 45.7 & 52.6 \\ 
Testra~\cite{zhao2022real} & 66.2 & 63.5 & 60.5 & 57.4 & 54.8 & 52.6 & 50.5 & 48.9 & 56.8 \\
MAT~\cite{wang2023memory} & - & - & - & - & - & - & - & - & 58.2 \\ 
\CMeRT\ & \textbf{69.9} & \textbf{66.6} & \textbf{63.2} & \textbf{60.1} & \textbf{57.3} & \textbf{54.9} & \textbf{52.8} & \textbf{50.9} & \textbf{59.5} \\
\hline
\end{tabular}
\vspace{-1mm}
\end{table}

\subsection{State-of-the-art Comparisons}

\subsubsection{Online Action Detection}
We compare \CMeRT\ with the state-of-the-art OAD methods on THUMOS'14, CrossTask and EK100 in \cref{tab: oad}. On THUMOS'14, our method outperforms memory-based methods LSTR~\cite{xu2021long}, Testra~\cite{zhao2022real}, and MAT~\cite{wang2023memory} by 3.7\%, 2\%, and 1.6\% in mAP, respectively. It achieves state-of-the-art performance, surpassing the latest work~\cite{guermal2024joadaa} by 0.6\% mAP. On CrossTask and EK100, our method also achieves state-of-the-art performance, improving mAP by 2\% on CrossTask and Top-5 action recall by 1.3\% on EK100. 

We further test naive baselines to address the train-inference discrepancy, including a reweighting method, MAT-rw and a streaming training method, MAT-stream, based on the state-of-the-art model MAT~\cite{wang2023memory}. For MAT-rw, a larger weight is assigned to the latest frame's loss. As learning in MAT is biased toward intermediate frames(\cref{fig:motivation}(b)), reweighting the latest frame helps mitigate the bias, resulting in slight performance improvements. For MAT-stream, only the latest frame in the short-term is used for training, discarding other short-term frames to match inference. We adjust the batch size accordingly to ensure that MAT-stream receives the same number of updates each epoch as MAT. Results indicate a large performance drop, possibly due to increased batch diversity. Previously, a training batch contains frames from the same short-term memory with similar views and actions. Removing all but the latest frame increases diversity within the batch, complicating training. In addition, MAT-stream also increases the training cost, as an entire forward pass is needed for training on a single frame.

\begin{table}
\centering
\vspace{-1mm}
\caption{Action anticipation results on EK100, measured by class mean Top-5 Recall.}
\label{tab: aa_ek100}
\vspace{-2mm}
\setlength{\tabcolsep}{1.5mm}
\small
\begin{tabular}{ccccc}
\hline
Method & Pre-train & Verb & Noun & Action\\ 
\hline
RULSTM~\cite{furnari2019would} & IN-1K & 27.8 & 30.8 & 14.0 \\
TempAgg~\cite{sener2020temporal} & IN-1K & 23.2 & 31.4 & 14.7 \\
AVT~\cite{girdhar2021anticipative} & IN-21K & 28.2 & 32.0 & 15.9 \\
Testra~\cite{zhao2022real} & IN-1K & 30.8 & 35.8 & 17.6 \\
MAT~\cite{wang2023memory} & IN-1K & 35.0 & 38.8 & 19.5 \\ 
\CMeRT\ & IN-1K & \textbf{35.1} & \textbf{39.7} & \textbf{19.8} \\
\hline
\end{tabular}
\vspace{-2mm}
\end{table}

\begin{table}[t]
\caption{Ablation study of Context-Enhancement(CE) using near-past and Memory-Refinement(MR) using near-future in OAD.}
\vspace{-2mm}
\centering
\small
\setlength{\tabcolsep}{1.5mm}
\begin{tabular}{ccccccc}
\hline
\multirow{2}{*}{CE} & \multirow{2}{*}{MR} & \multirow{2}{*}{TH'14} & \multirow{2}{*}{CrossTask} & \multicolumn{3}{c}{EK100} \\ 
\cline{5-7} & & & & Verb & Noun & Action \\ \hline
\color{Red}\xmark & \color{Red}\xmark & 71.5 & 33.4 & 44.9 & 26.9 & 26.3\\ 
\color{Red}\xmark & \color{Green}\cmark & 73.0 & 34.8 & 46.7 & 47.3 & 27.1 \\ 
\color{Green}\cmark & \color{Red}\xmark & 71.9 & 33.9 & 46.0 & 47.6 & 26.6\\ 
\color{Green}\cmark & \color{Green}\cmark & \textbf{73.2} & \textbf{35.9} & \textbf{47.1} & \textbf{48.3} & \textbf{27.6} \\ 
\hline
\end{tabular}
\label{tab:ab_comb}
\vspace{-5mm}
\end{table}

\subsubsection{Action Anticipation}
We compare our method to prior approaches on THUMOS'14 and EK100 for action anticipation in \cref{tab: aa_th} and \cref{tab: aa_ek100}. Our method outperforms the state-of-the-art MAT~\cite{wang2023memory} by 1.3\% on THUMOS'14 and achieves competitive performance on EK100, particularly in noun and action prediction, surpassing MAT by 0.9\% and 0.3\%, despite the dataset's large scale and diverse categories. The results validate our method's effectiveness in jointly modeling detection and anticipation. Anticipation can be further improved, but at the cost of detection due to the inherent task trade-off.

\subsection{Ablation Studies}
\noindent \textbf{Contribution of Key Modules.} We assess the contribution of the near-past context in the Context-Enhanced encoder and the near-future generation in the Memory-Refined decoder in \cref{tab:ab_comb}. Incorporating near-future generation for refinement improves performance by 1.5\%, 1.4\%, and 0.8\% on THUMOS'14, CrossTask, and EK100, respectively. Context enhancement further boosts performance by approximately 0.2\% on THUMOS'14, 1.1\% on CrossTask, and 0.5\% on EK100. Combined, these two modules yield the best results.

\noindent \textbf{Near-Past Context. } The near-past context is introduced to enrich short-term memory with additional past context, especially for earlier frames. Results in \cref{tab:ab_ce} show the impact of near-past context length, indicating that a moderate length relative to short-term memory is optimal. The near-past length balances between enhancing early frame representations with more near-past context and providing data augmentation with less. A longer near-past length may lead to overfitting of the lateset frame by giving it excessive context(both short-term and near-past context), while no near-past context results in hard training samples(these earlier frames), offering context-based data augmentation. For THUMOS'14, a shorter length of 0.5 seconds performs best, likely due to the dataset's simplicity and overfitting risk. In contrast, CrossTask and EK100 benefit from a longer near-past length to capture complex action interactions.

\begin{table}
\centering
\caption{Ablation study of near-past(N-past) length(sec.) for Context-Enhanced Encoder in OAD.}
\label{tab:ab_ce}
\vspace{-2mm}
\small
\setlength{\tabcolsep}{1.3mm}
\begin{tabular}{ccccccccc}
\cline{1-2} \cline{4-8}
\multirow{2}{*}{N-past} & \multirow{2}{*}{CrossTask} & \multirow{2}{*}{ } & \multirow{2}{*}{N-past} & \multirow{2}{*}{TH'14} &\multicolumn{3}{c}{EK100} \\ 
\cline{6-8} & & & & & Verb & Noun & Action \\ 
\cline{1-2} \cline{4-8}
5 & 35.1 & & 0.5 & \textbf{73.2} & 46.4 & 47.7 & 27.2\\ 
10 & \textbf{35.9} & & 1 & 72.8 & 46.8 & 47.6 & 27.3 \\ 
15 & 35.6 & & 2 & 72.7 & \textbf{47.1} & \textbf{48.3} & \textbf{27.6}\\ 
20 & 35.5 & & 3 & 72.4 & 46.7 & 47.9 & 27.5 \\
\cline{1-2} \cline{4-8}
\end{tabular}
\vspace{-2mm}
\end{table}

\begin{table}
\centering
\caption{Ablation study on near-future(N-fut) length(sec.) for Memory-Refinement module in OAD.}
\label{tab:ab_fr}
\vspace{-2mm}
\small
\setlength{\tabcolsep}{1.7mm}
\begin{tabular}{cccccc}
\hline
\multirow{2}{*}{N-fut(s)} & \multirow{2}{*}{TH'14} & \multirow{2}{*}{CrossTask} & \multicolumn{3}{c}{EK100} \\ 
\cline{4-6} & & & Verb & Noun & Action \\ 
\hline
4 & 72.6 & 35.1 & 45.9 & 48.5 & 27.0\\ 
8 & 73.0 & 35.3 & \textbf{47.1} & 48.3 & \textbf{27.6} \\ 
12 & \textbf{73.2} & \textbf{35.9} & 46.7 & \textbf{48.6} & 27.3 \\ 
16 & 72.7 & 35.6 & 46.8 & 27.6 & 27.1 \\ 
\hline
\end{tabular}
\vspace{-4mm}
\end{table}

\begin{table}[b]
\vspace{-4mm}
\caption{Ablation study on near vs. distant contexts in OAD.}
\vspace{-2mm}
\centering
\small
\setlength{\tabcolsep}{1.5mm}
\begin{tabular}{ccccccc}
\hline
\multirow{2}{*}{Past} & \multirow{2}{*}{Future} & \multirow{2}{*}{TH'14} & \multirow{2}{*}{CrossTask} & \multicolumn{3}{c}{EK100} \\ 
\cline{5-7} & & & & Verb & Noun & Action \\ \hline
Near & Near & 73.2 & 35.9 & 47.1 & 48.3 & 27.6\\ 
Near & Distant & 72.9 & 35.0 & 46.6 & 47.8 & 27.0 \\ 
Distant & Near & 73.0 & 35.3 & 46.8 &48.0 & 27.1\\ 
\hline
\end{tabular}
\label{tab:ab_nd}
\vspace{-4mm}
\end{table}

\begin{table*}[ht!]
\begin{minipage}[t]{0.33\textwidth}
\caption{Efficiency comparison on TH'14.}
\label{tab:ab_speed}
\centering
\vspace{-2mm}
\resizebox{1.0\columnwidth}{!}{
\setlength{\tabcolsep}{0.6mm}
\renewcommand{\arraystretch}{1.36}
\begin{tabular}{ccccc}
\hline
Method & \#Param & GFLOPS & FPS & mAP \\ 
\hline
LSTR~\cite{xu2021long} & 58.8 & 4.70 & 140.8 & 69.5 \\ 
Testra~\cite{zhao2022real} & 58.9 & 4.72 & 135.1 & 71.2 \\ 
MAT~\cite{wang2023memory} & 107.4 & 6.62 & 102.0 & 71.6 \\ 
\CMeRT\ & 94.5 & 5.36 & 126.6 & 73.2 \\ 
\hline
\end{tabular}
}
\end{minipage}
\
\begin{minipage}[t]{0.29\textwidth}
\caption{Performance using DinoV2. }
\label{tab:ab_feat}
\centering
\vspace{-2mm}
\resizebox{1.0\columnwidth}{!}{
\setlength{\tabcolsep}{0.9mm}
\renewcommand{\arraystretch}{1.18}
\begin{tabular}{ccccc}
\hline
\multirow{2}{*}{Method} & \multicolumn{2}{c}{TH'14} & \multicolumn{2}{c}{CrossTask} \\
\cline{2-5}
 & ResNet & Dino & I3D & Dino \\ 
\hline
LSTR~\cite{xu2021long} & 69.5 & 74.3 & 33.0 & 45.1 \\ 
Testra~\cite{zhao2022real} & 71.2 & 74.5 & 33.4 & 44.9 \\ 
MAT~\cite{wang2023memory} & 71.6 & 75.3 & 33.9 & 46.8 \\ 
\CMeRT\ & 73.2 & 76.4 & 35.9 & 47.3 \\ 
\hline
\end{tabular}
}
\end{minipage}
\
\begin{minipage}[t]{0.35\textwidth}
\caption{Online action detection with latency $\delta$.}
\label{tab:oad_lat}
\centering
\vspace{-2mm}
\small
\resizebox{1.0 \columnwidth}{!}{
\setlength{\tabcolsep}{0.8mm}
\begin{tabular}{cccccc}
\hline
\multirow{2}{*}{Latency} & \multirow{2}{*}{TH'14} & \multirow{2}{*}{CrossTask} & \multicolumn{3}{c}{EK100} \\ 
\cline{4-6} & & & Verb & Noun & Action \\ \hline
 0s & 73.2 & 35.9 & 47.1 & 48.3 & 27.6 \\
0.25s & 74.5 & - & 48.2 & 49.0 & 27.9 \\
0.5s & 75.4 & - & 48.9 &49.5 & 28.2\\
1s & 76.6 & 36.6 & 49.4 & 50.6 & 28.7 \\
2s & 76.7 & 36.9 & 49.7 & 50.4 & 29.3 \\
\hline
\end{tabular}}
\end{minipage}%
\vspace{-0mm}
\end{table*}

\begin{table*}[ht!]
\caption{Benchamarking results with new protocols, including frame-wise(mAP and Rec@5) and event-wise(point-wise F1 score with threshold 1s(P-F1) and segment-wise F1 score with iou threshold 0.25(S-F1)) metrics, updated features, and latency models. For EK100, we report frame-wise performance for verb/noun/action(v/n/a), and event-wise performance only on actions.}
\centering
\vspace{-2mm}
\small
\setlength{\tabcolsep}{1.5mm}
\begin{tabular}{cccccccccccccc}
\hline
\multirow{2}{*}{Method} & \multicolumn{4}{c}{THUMOS'14} & \multicolumn{4}{c}{CrossTask} & \multicolumn{5}{c}{EK100} \\ 
\cline{2-14} & mAP & P-F1 & Edit & S-F1 & mAP & P-F1 & Edit & S-F1 & mAP(v/n/a) & Rec@5(v/n/a) & P-F1 & Edit & S-F1 \\ 
\hline
LSTR~\cite{xu2021long} & 70.0 & 42.9 & 45.9 & 49.0 & 33.0 & 26.2 & 34.2 & 34.8 & 15.6/16.0/8.6 & 39.1/44.3/23.5 & 7.1 & 7.2 & 6.0 \\ 
Testra~\cite{zhao2022real} & 71.2 & 42.7 & 44.4 & 47.3 & 33.4 & 25.0 & 34.0 & 33.7 & 16.4/18.3/9.8 & 41.1/45.8/25.1 & 8.1 & 8.0 & 7.2 \\ 
MAT~\cite{wang2023memory} & 71.6 & 45.0 & 45.1 & 50.0 & 33.9 & 26.5 & 35.5 & 34.4 & 16.4/18.9/10.8 & 43.5/46.9/26.3 & 8.6 & 9.1 & 7.8 \\ 
\CMeRT\ & \textbf{73.2} & \textbf{45.8} & \textbf{46.9} & \textbf{51.5} & \textbf{35.9} & \textbf{28.4} & \textbf{36.8} & \textbf{37.0} & \textbf{18.5}/\textbf{19.7}/\textbf{11.5} & \textbf{47.0}/\textbf{48.6}/\textbf{27.6} & \textbf{10.7} & \textbf{10.9} & \textbf{10.4} \\ 
\hline 
\CMeRT\ Dinov2 & 76.4 & 47.8 & 49.1 & 52.8 & 47.3 & 35.4 & 44.5 & 44.9 & - & - & - & - & - \\
\CMeRT\ latency@1s & 76.6 & 48.6 & 55.2 & 55.6 & 36.6 & 29.1 & 38.1 & 39.4 & 19.8/20.8/12.2 & 49.4/50.6/28.7 & 13.2 & 13.9 & 14.1 \\ 
\hline 
\end{tabular}
\label{tab:new_metrics_all}
\vspace{-4mm}
\end{table*}

\noindent \textbf{Near-Future Context. }
The generated near-future enriches short-term memory with information beyond the past, enhancing detection. \cref{tab:ab_fr} shows the impact of near-future length, suggesting that an optimal length should exceed short-term memory, enabling all short-term frames to access immediate future context. A length that’s too short fails to provide future information for the latest frames, while an overly long length makes long-horizon prediction challenging, potentially distracting detection learning and introducing noise.

\noindent \textbf{Near \emph{vs.} Distant Context. } To highlight the importance of immediate context, we conduct ablation studies by replacing the near-past($t_s-T_c$ to $t_s$) and near-future($t_s$ to $t_s + T_f$) contexts with their distant counterparts. Specifically, we use distant-past from $t_s - T_c - T_s$ to $t_s - T_s$ and distant-future from $t$ to $t + T_f$ . The results in \cref{tab:ab_nd}, demonstrate the importance of near-contexts over distant ones. Especially, distant future leads to a larger performance drop, as long-horizon anticipation is harder than the near one.

\subsection{Runtime Analysis}
We analyze computational complexity and runtime speed using a single NVIDIA RTX A4000 GPU on THUMOS’14. Focusing on runtime excluding preprocessing (i.e., without feature extraction), \cref{tab:ab_speed} compares model parameters, computational complexity, FPS, and performance. Compared to LSTR~\cite{xu2021long} and Testra~\cite{zhao2022real}, our model is larger and has higher computational cost due to near-future generation and memory refinement, resulting in lower FPS but yielding large performance gains. Against the SOTA MAT~\cite{wang2023memory}, which also uses generated future, our method is more efficient, achieving 24.6 FPS higher with a smaller model and reduced computational cost, while also delivering superior performance. Like \cite{xu2021long}, our method can further enhance online inference efficiency by storing intermediate results from the first compression stage, increasing FPS to 133.3. 

\subsection{Advancing OAD}
\noindent \textbf{State-of-the-art Features.} Existing OAD works rely on outdated pre-extracted RGB features, \eg ResNet or segment-wise features, \eg I3D. Given advancements in vision foundation models~\cite{oquab2023dinov2, zhai2023sigmoid}, we explore using the advanced DinoV2~\cite{oquab2023dinov2} for feature extraction, evaluated on THUMOS'14 and CrossTask. As shown in \cref{tab:ab_feat}, DinoV2 features yield performance gains of 3.2\% and 11.4\% on THUMOS'14 and CrossTask, respectively. Testing all memory-based methods with new features further confirms our method's robustness, as it consistently outperforms others. Advanced features have promising implications for OAD research community, suggesting that models are not always the bottleneck; advancing the field requires consistently using state-of-the-art features that capture fine-grained details.

\noindent \textbf{Sequence Metrics.} While current methods are mainly evaluated frame-wise, understanding event-wise performance is equally crucial for OAD. High frame-wise accuracy can mask low event-wise accuracy, especially for short-duration actions and cases of oversegmentation~\cite{farha2019ms, pang2024long}. We encourage using event-based metrics inspired by start-point detection~\cite{gao2021woad, gao2019startnet} and temporal action segmentation~\cite{farha2019ms, singhania2023c2f, yi2021asformer, pang2024cost}: the Point-wise F1 score (1s threshold), Segment-wise F1 score (IoU threshold of 0.25), and the Edit score. Benchmarking memory-based methods in \cref{tab:new_metrics_all}, \CMeRT\ consistently outperforms others on both frame-wise and event-wise metrics, highlighting its superior robustness.

\noindent \textbf{OAD with Latency.} Observing the performance gains from indirect future access in intermediate short-term frames in \cref{subsec: leak}, we explore the effect of directly accessing the future frames by introducing the future latency $\delta$. Future latency $\delta$ enables predictions for frame $t-\delta$ at time $t$, providing a preview of the near future. It is valuable for applications that can tolerate delays or require post-prediction refinement. We introduce the first OAD baseline with future latency by replacing the causal mask with a new latency mask(see details in Suppl. C), enabling each short-term frame to access both past and near-future information up to a limit of $\delta$. This immediate future context is essential, markedly boosting detection performance, as shown in \cref{tab:oad_lat}. Even a slight latency, \eg $\delta=0.25s$\footnote{Features are extracted at 4 FPS, with average action durations of 4.9s(THUMOS'14) and 3.3s(EK100). A 0.25s delay equals to just 1 frames.} can lead to greater improvements.

\subsection{Qualitative results} 
\begin{figure}[htb]
\centering
\vspace{0mm}
\includegraphics[width=0.99\linewidth]{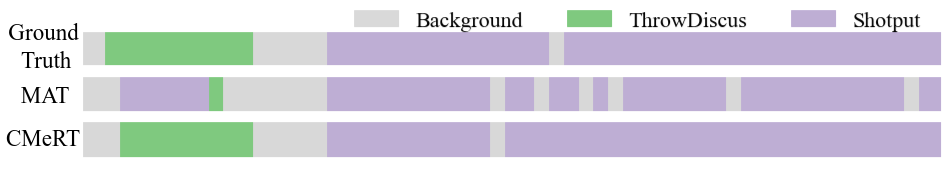} 
\vspace{-3mm}
\caption{Quality results on THUMOS'14}
\label{fig: quality}
\vspace{-3mm}
\end{figure}

\cref{fig: quality} shows the qualitative results for THUMOS'14. The bar charts compare the ground truth, predictions from MAT~\cite{wang2023memory} and our method \CMeRT. Our method addresses learning biases, leading to more robust representations that better separate actions from the background or similar actions~\textit{(ThrowDiscus vs. Shotput)}. But, it struggles with short actions or small subjects in similar backgrounds(see more results in Supplementary D). These issues can be mitigated using advanced features that capture finer details. 

%% file: sec/6_conclusion.tex
\section{Conclusion}
\label{sec:conclusion}

This paper identifies the training-inference discrepancy in memory-based OAD methods, resulting in poor frame representations and learning biases in short-term memory. We propose the Context-Enhanced Memory-Refined Transformer(\CMeRT) for joint detection and anticipation. \CMeRT\ integrates near-past and near-future to provide additional immediate context, ensuring consistent learning across short-term frames. We also introduce a new OAD benchmark and protocols to
better align with practical applications.
Future efforts will explore trade-offs between detection and anticipation within the unified framework, and leveraging prior knowledge to capture high-level temporal dependencies.

%% file: sec/X_suppl.tex
\clearpage
\setcounter{page}{1}
\maketitlesupplementary

\section*{A. Diagnising Context Modeling for OAD}
\begin{figure}[ht!]
     \centering
     \begin{minipage}[t]{0.4\textwidth}
     \includegraphics[width=0.98\linewidth]{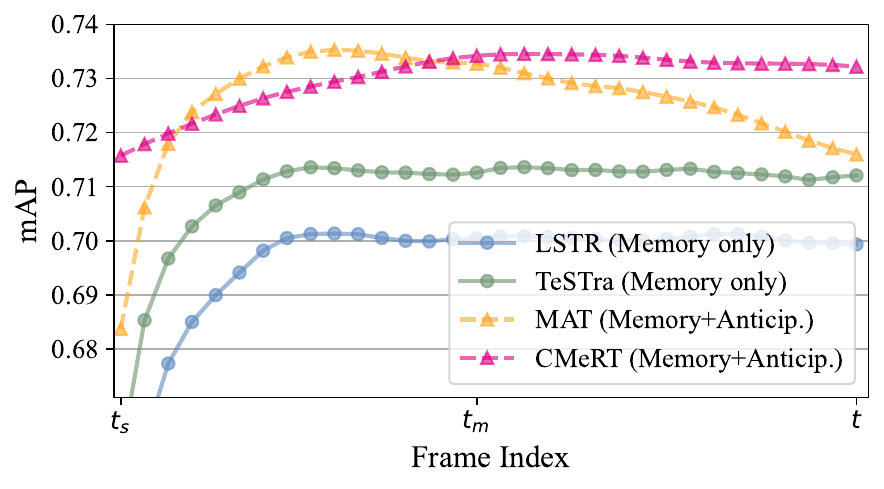}
     \vspace{-8mm}
    \vspace{-3mm}
    \end{minipage}
  \hfill
     \begin{minipage}[t]{0.4\textwidth}
     \includegraphics[width=0.985\linewidth]{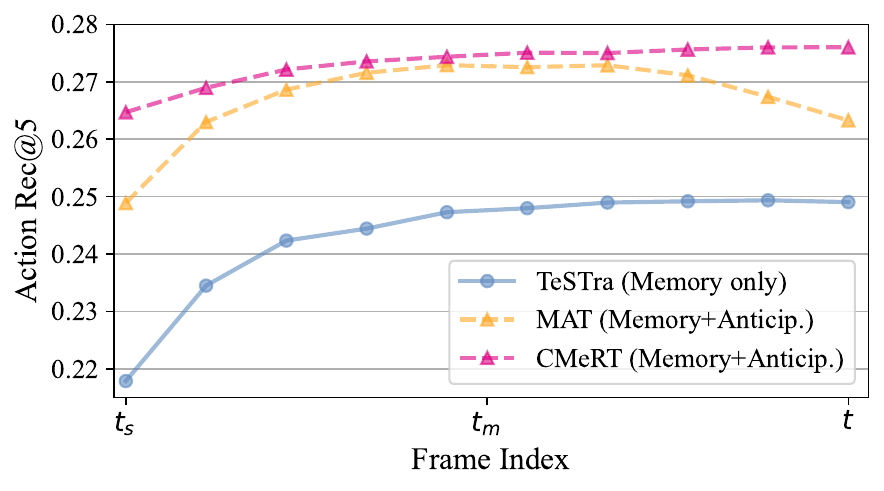}
     \vspace{-8mm}
    \end{minipage}
  \vspace{-3mm}
  \caption{Frame performance within the short-term memory on THUMOS'14(top) and EK100(bottom).}
  \label{fig: supp_map}
  \vspace{-3mm}
\end{figure}

Existing methods~\cite{xu2021long, zhao2022real, wang2023memory} suffer from a training-inference discrepancy, causing short-term context imbalance and a non-causal leakage during anticipation, resulting in learning biases. \cref{fig: supp_map} shows the learning biases present in existing works from a performance perspective. 
\vspace{-3mm}

First, we observe that early frames in the short-term memory are poorly learned, resulting in significantly lower performance. These poorly represented frames serve as low-quality samples, impairing the learning of classifier to effectively predict the latest frame. In contrast, \CMeRT\ improves performance for early frames, though a performance gap still remains compared to the latest frame. 
The remaining gap is due to the use of a shorter near-past context, which limits the amount of context available to earlier frames compared to the latest one. Our empirical findings demonstrate that shorter near-past contexts are more beneficial, as they act as a form of data augmentation by exposing frames to less context.  Naive approaches~\cite{xu2021long, zhao2022real, wang2023memory} that omit near-past context can also be seen as a form of data augmentation. However, they over-augment the data, introducing poor training samples that hamper the learning process.

Second, the performance curve of the anticipation-based method MAT~\cite{wang2023memory} confirms the presence of non-causal leakage, as it shows significantly higher performance for intermediate frames compared to the latest frame. \CMeRT\, however, effectively mitigate this leakage and learning bias, prioritizing the learning of the latest frame.

\section*{B. Experiments}
\paragraph{Hyperparameters. } The hyperparameters used for each dataset are summarized in \cref{tab:supp_hyp}.

\begin{table}[th!]
\vspace{-2mm}
\caption{Hyperparameters for different experimental settings.}
\centering
\vspace{-2mm}
\small
\resizebox{0.9\columnwidth}{!}{
\begin{tabular}{cccc}
\hline
  & THUMOS'14 & CrossTask & EK100 \\ 
\hline
batch size & 32 & 32 & 32 \\ 
epoch & 12 & 12 & 12 \\
warmup & 8 & 5 & 10 \\
learning rate &2e-4 & 7e-5 & 7e-5 \\
weight decay & 5e-5 & 1e-5 & 1e-4 \\
\hline
\end{tabular}}
\label{tab:supp_hyp}
\vspace{-8mm}
\end{table}

\paragraph{MAT-rw and MAT-stream.}
We implement MAT-rw and MAT-stream based on the state-of-the-art memory-based model MAT~\cite{wang2023memory} to evaluate standard approaches for addressing the training-inference discrepancy. 

In MAT-rw, we assign a higher weight to the loss of the latest frame to mitigate the learning bias towards intermediate frames. Specifically, the weight is set to 1.2 for THUMOS’14 and 3.0 for CrossTask and EK100.

In MAT-stream, only the latest frame in the short-term memory is used for training, while other short-term frames are discarded to align with the inference. we modify the sliding window sampling by setting the stride to 1, ensuring all video frames are used for training. However, this increases the training set size compared to using a stride equal to the short-term memory length, resulting in more training samples and updates per epoch than the standard MAT. To mitigate this, we adjust the batch size to match the number of updates per epoch as in MAT~\cite{wang2023memory}.

\section*{C. Advancing OAD}
\paragraph{DinoV2 Features.} We use the Dinov2 ViT-g/14 model~\cite{oquab2023dinov2} to extract advanced RGB features for THUMOS'14 and CrossTask. We replace only the RGB features while other features, such as optical flow, remain unchanged. For THUMOS'14, following \cite{xu2021long}, we extract video frames at a rate of 24 FPS and divide the video into chunks of 6 frames, using the intermediate frame of each chunk for RGB feature extraction. The feature extraction is performed at the chunk level, meaning evaluation occurs every 0.25 seconds. 
The feature encoding process for CrossTask is similar to THUMOS"14, except that the chunk size is increased to 24 frames to align with the existing feature set. 

While the advanced feature extractor improves performance, it also increases the computational burden. Following \cite{xu2021long}, we report the runtime for end-to-end online inference on THUMOS'14, including two-stream feature extraction in \cref{tab:ab_speed1}. Specifically, DenseFlow~\cite{densflow} is used to compute optical flow, while RGB features are extracted using either ResNet52~\cite{wang2016temporal} or the DinoV2 model. The results in \cref{tab:ab_speed1} align with prior works~\cite{wang2023memory, xu2021long}, confirming that optical flow remains the primary speed bottleneck. Compared to optical flow feature, the runtime for DinoV2 RGB feature extraction remains manageable. However, the DinoV2 model inference can be further accelerated through techniques such as model distillation, model weights quantization or conversion to Optimized formats, like TorchScript and ONNX. Model inference optimization is already a well-established practice in the industry, providing significant opportunities to leverage more advanced features while maintaining efficiency.

\begin{table}[th]
\vspace{-2mm}
\caption{Efficiency analysis of feature extraction on THUMOS'14. The performance is reported in frames per second(FPS) }
\centering
\vspace{-2mm}
\resizebox{0.85\columnwidth}{!}{
\begin{tabular}{cc|cc}
\hline
\multicolumn{2}{c|}{\textbf{Optical Flow}} & \multicolumn{2}{c}{\textbf{RGB}} \\ 
\hline
Computation & Extraction & ResNet52 & DinoV2 \\
\hline
8.6 & 47.6 & 69.0 & 13.9  \\ 
\hline
\end{tabular}}
\label{tab:ab_speed1}
\vspace{-3mm}
\end{table}

\paragraph{OAD with latency}

\begin{figure}[ht!]
\vspace{-7mm}
\centering
\includegraphics[width=0.95\linewidth]{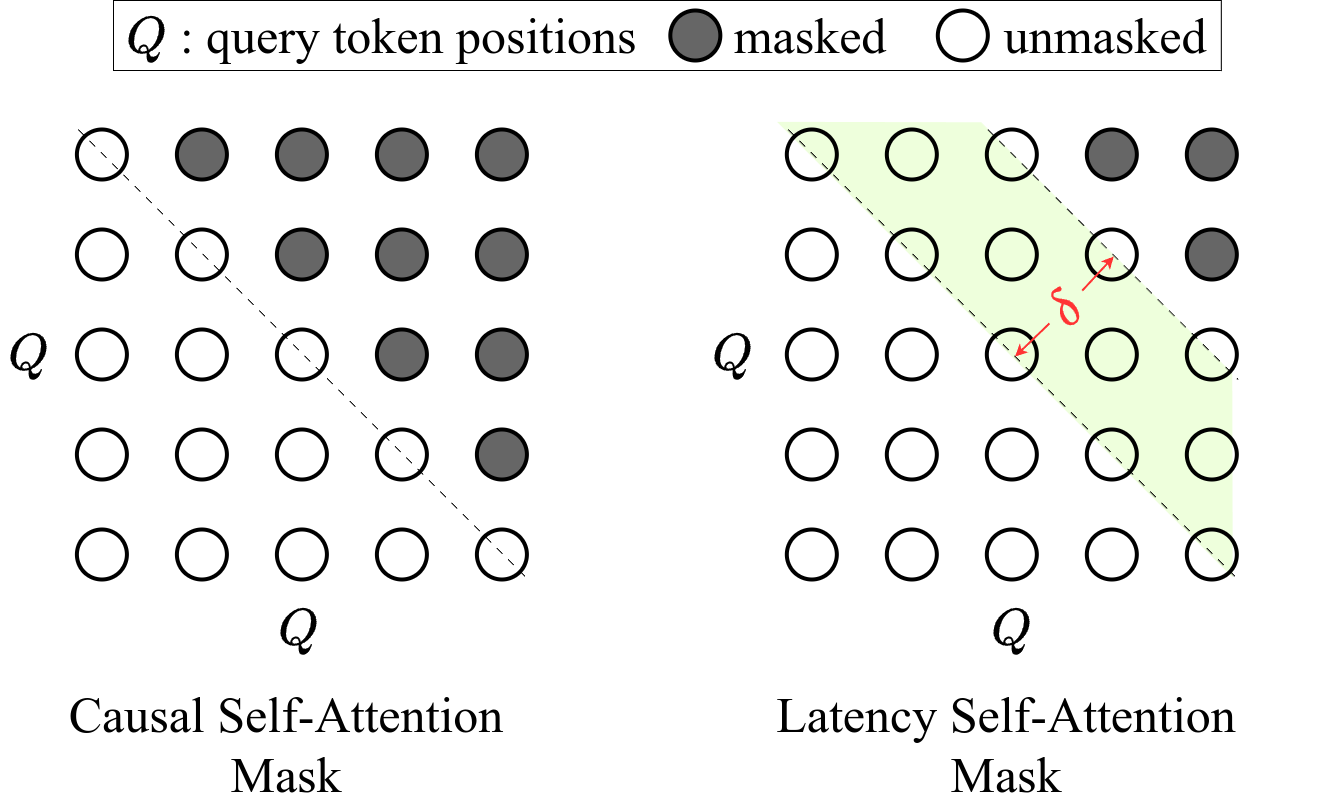}
\vspace{-2mm}
\caption{Self-attention masking to control query interactions.}
\vspace{-2mm}
\label{fig: supp_mask}
\end{figure}

For applications where delays are acceptable or post-prediction refinement is required, it is valuable to explore the advantages of incorporating limited future information into online action detection. To explore this, we introduce a future latency parameter, $\delta$, and propose the first OAD baseline with future latency. Specifically, we construct base models based on Testra~\cite{zhao2022real}, MAT~\cite{wang2023memory}, and our model \CMeRT\ by replacing the causal mask in short-term self-attention with a new latency mask, as shown in \cref{fig: supp_mask}. This new mask allows each short-term frame to additionally access the near-future information up to a limit of $\delta$. 

\begin{figure}[ht]
     \centering
     \begin{minipage}[t]{0.365\textwidth}
     \includegraphics[width=1.0\linewidth]{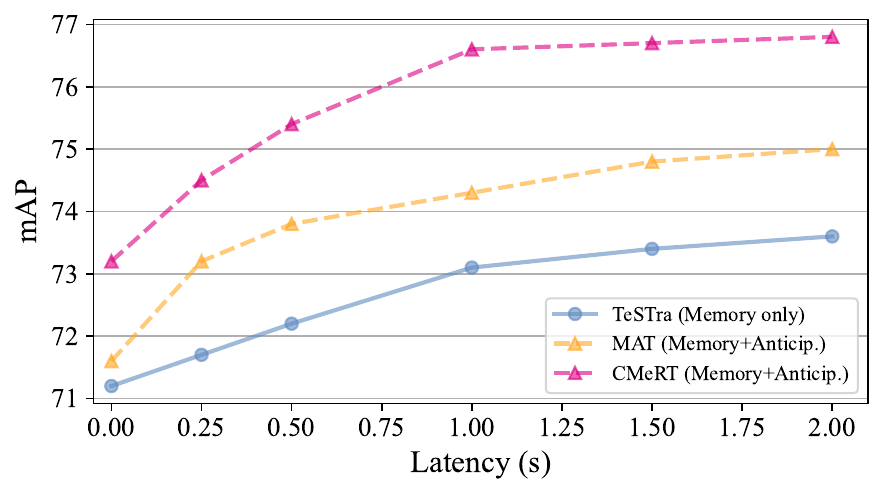}
     \vspace{-8mm}
    \vspace{-1mm}
    \end{minipage}
  \hfill
     \begin{minipage}[t]{0.365\textwidth}
     \includegraphics[width=1.0\linewidth]{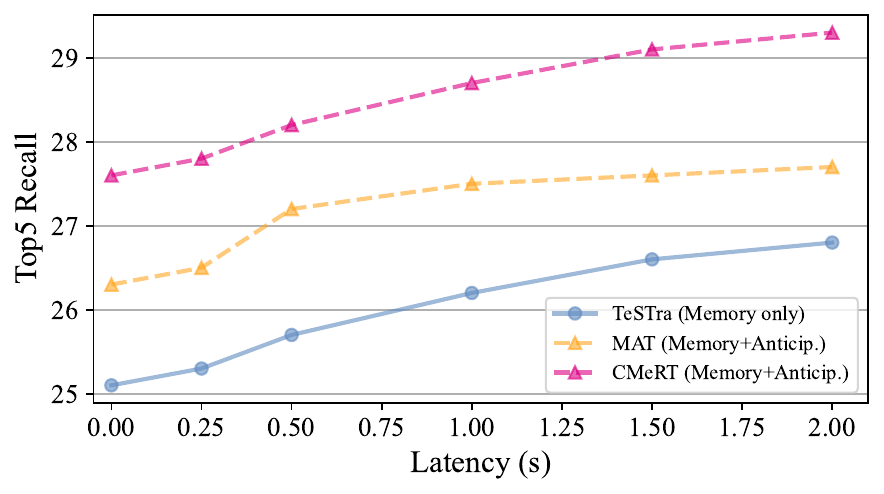}
     \vspace{-8mm}
    \vspace{-3mm}
    \end{minipage}
  \vspace{-3mm}
  \caption{OAD performance under varied future latency on THUMOS'14(top) and EK100(bottom).}
  \label{fig: supp_lat}
  \vspace{-6mm}
\end{figure}

\begin{figure*}[htb]
\centering
\includegraphics[width=1.0\linewidth]{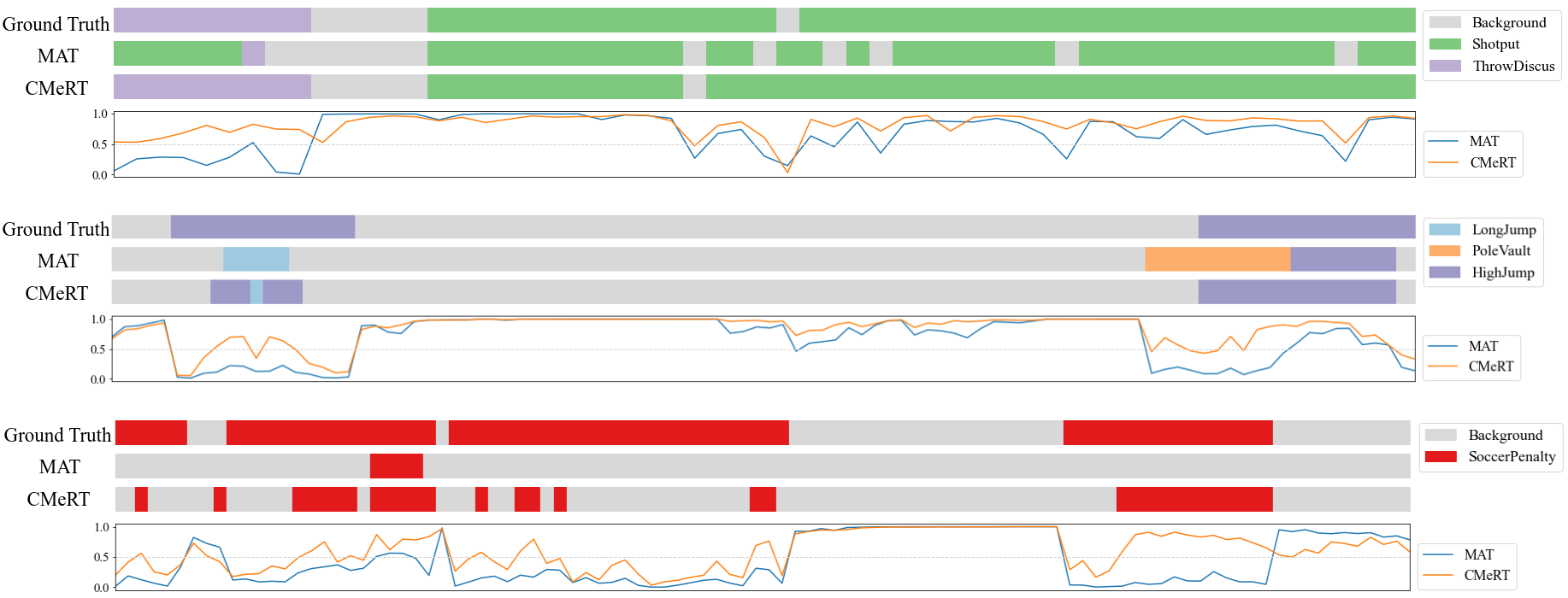}
\vspace{-4mm}
\caption{Quality results on THUMOS'14 - bar charts show predictions; curve plots for confidence of the true action.}
\label{fig: supp_quality}
\end{figure*}

\begin{figure*}[htb]
\centering
\includegraphics[width=1.0\linewidth]{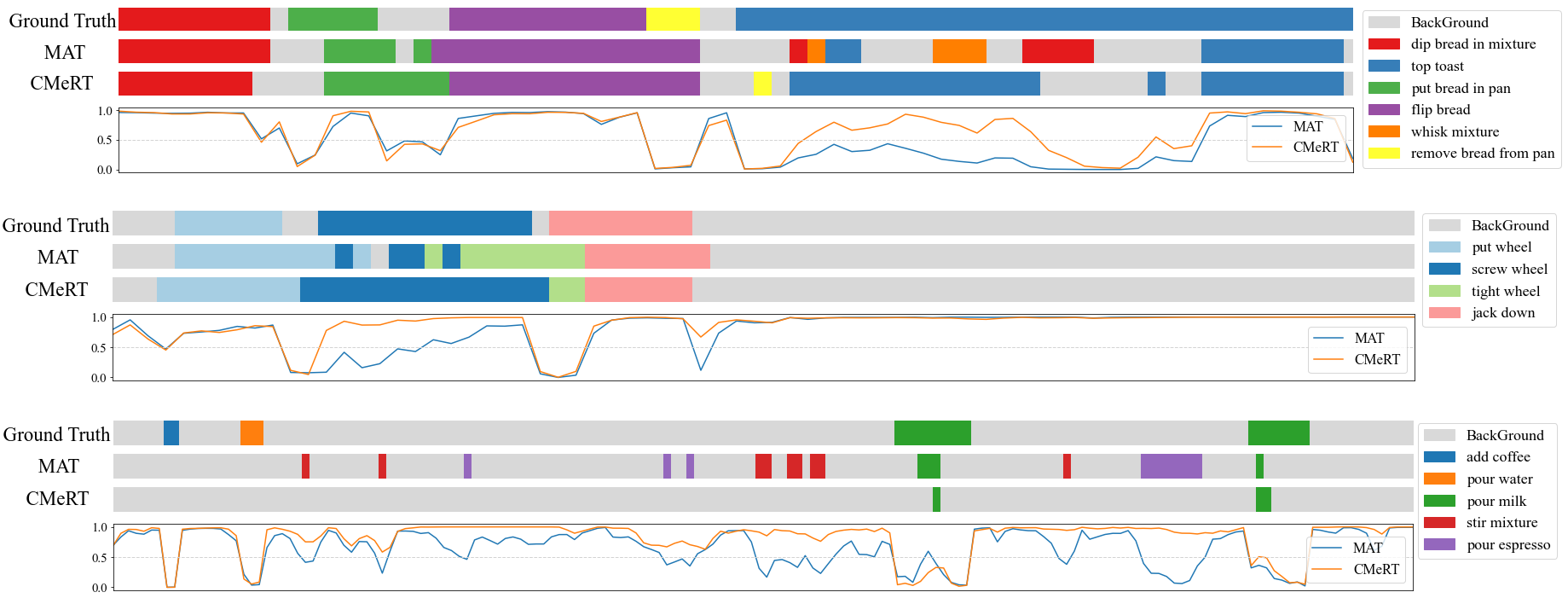}
\vspace{-4mm}
\caption{Quality results on CrossTask: top - Make French Toast,  middle - Change a Tire, bottom - Make a Latte}
\label{fig: supp_quality1}
\end{figure*}

We evaluate the new OAD with latency setting using various base models and latency settings, with results in \cref{fig: supp_lat}. Incorporating future latency improves performance across all models. Even a small latency, \eg $\delta=0.5$ can lead to greater improvements, with further gains expected as the latency increases. \CMeRT\ consistently outperforms others by a large margin, demonstrating its robustness.

\section*{D. Qualitative Results}
\cref{fig: supp_quality} and \cref{fig: supp_quality1} show some qualitative results for THUMOS'14 and CrossTask, respectively.
The bar charts present a comparison between the ground truth and the predictions from MAT~\cite{wang2023memory} and our method \CMeRT. The curve plots display the confidence in identifying the current true action. The results highlight that \CMeRT\ effectively reduce the misclassification between background and foreground actions. Additionally, it improves the distinction between similar actions~\textit{(PoleVault vs. HighJump)}. However, it struggles with short actions~\textit{(Whisk mixture \& add coffee)} or small subjects in similar backgrounds~\textit{(SoccerPenalty)}.

\section*{E. Extra Ablation Studies}
\noindent \textbf{Query configuration in the long-term compressor:} We test on four query configurations (stage1-stage2): 16-16, 16-32, 32-32, and 32-64 on THUMOS'14. The mAP is 72.8\%, 73.2\%, 72.9\%, and 72.8\%, respectively.
The results suggest that intermediate configurations are optimal, as excessive queries introduce noise and redundancy, while too few causes the loss of valuable information.
\begin{figure*}[hb!]
\begin{minipage}[t]{0.32\textwidth}
\centering
 \includegraphics[width=1.01\linewidth]{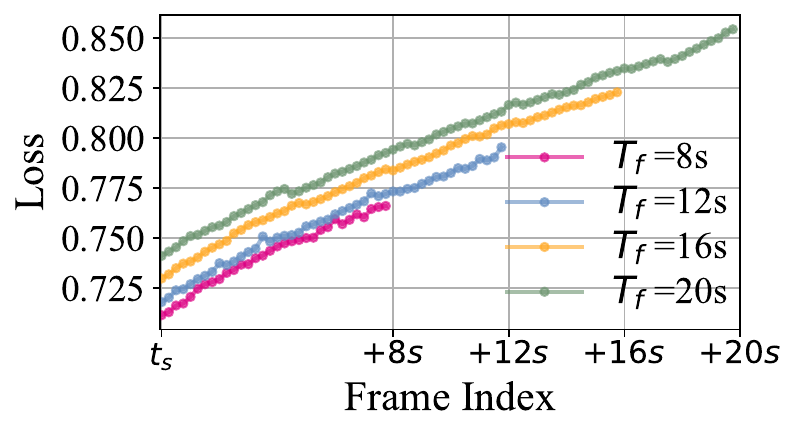}
\caption{Extended future generation reduces quality (on THUMOS'14). }
\label{fig:sup_dist1}
\end{minipage}
\
\begin{minipage}[t]{0.32\textwidth}
\centering
    \includegraphics[width=0.92\linewidth]{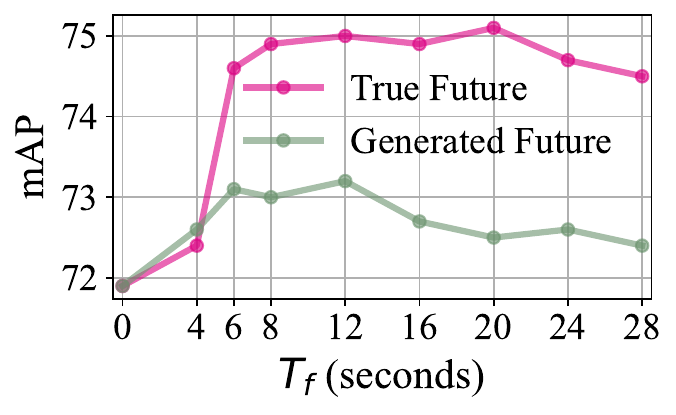}
\caption{Distant future not helpful (on THUMOS'14).}
\label{fig:sup_dist}
\end{minipage}
\
\begin{minipage}[t]{0.32\textwidth}
\centering
    \includegraphics[width=0.9\linewidth]{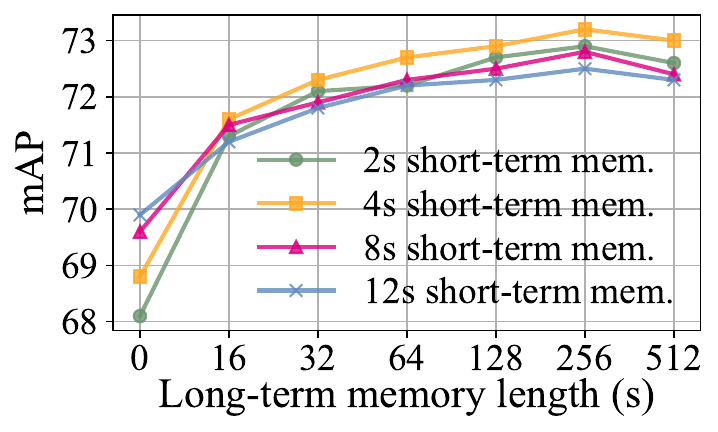}
\caption{Impact of long-short division on THUMOS'14.}
\label{fig:sup_lsd}
\end{minipage}
\vspace{-1mm}
\end{figure*}

\noindent \textbf{Short~over~long future:} We designed the long-term memory~($t_l$ to $t_s$) to generate a near-future~($t_s$ to $t_s + T_f$) that overlaps and extends beyond the short memory to serve a pseudo-future for all short-term frames. Experimentally,  generating a short near-future is favored over a longer one, as longer pseudo-futures are more challenging and costly, leading to degraded quality (\cref{fig:sup_dist1}). Even using the true future, performance saturates beyond a certain length (\cref{fig:sup_dist}), which justifies our use of short-future generation. 

\noindent  \textbf{Long-short division:} We evaluate the impact of long-short term division on performance. As shown in \cref{fig:sup_lsd}, excessive long-term memory introduces noise, while insufficient long-term causes information loss. The short-term length has minimal impact if sufficient long-term is provided. Besides, near-future generation is less impacted by the division, since it always predicts the future following the long-term.